%% file: main.tex
\documentclass[10pt,twocolumn,letterpaper]{article}

\usepackage{cvpr}              
\input{preamble}
\definecolor{cvprblue}{rgb}{0.21,0.49,0.74}
\usepackage[pagebackref,breaklinks,colorlinks,allcolors=cvprblue]{hyperref}
\usepackage[ruled]{algorithm2e}
\usepackage{tikz}
\usetikzlibrary{shapes, arrows, positioning, calc, patterns}
\usepackage{pgfplots}
\usepackage{multirow}
\pgfplotsset{compat=1.17}

\def\paperID{5054} 
\def\confName{CVPR}
\def\confYear{2026}

\title{InTrain: Intrinsic Trainability for Zero-Cost Neural Architecture Search}

\author{
    Qinqin Zhou$^{1}$  \quad Fuhai Chen$^{1}$ \quad Jipeng Wu$^{2}$ \quad Zhiwei Chen$^{3}$ \quad Zhikai Hu$^{4}$ \quad Weiwei Cai$^{5}$ 
    \and
    $^1$School of Computer and Data Science, Fuzhou University, \\
    $^2$School of Computer and Data Science, Minjiang University,\\
    $^3$School of Artificial Intelligence, Nanchang University,\\
    $^4$Department of Computer Science, Hong Kong Baptist University,\\
    $^5$School of Interdisciplinary Medicine and Engineering, Harbin Medical University
}

\begin{document}
\maketitle
\input{sec/0_abstract} 
\begin{figure}[t]
\centering
\includegraphics[width=0.4\textwidth]{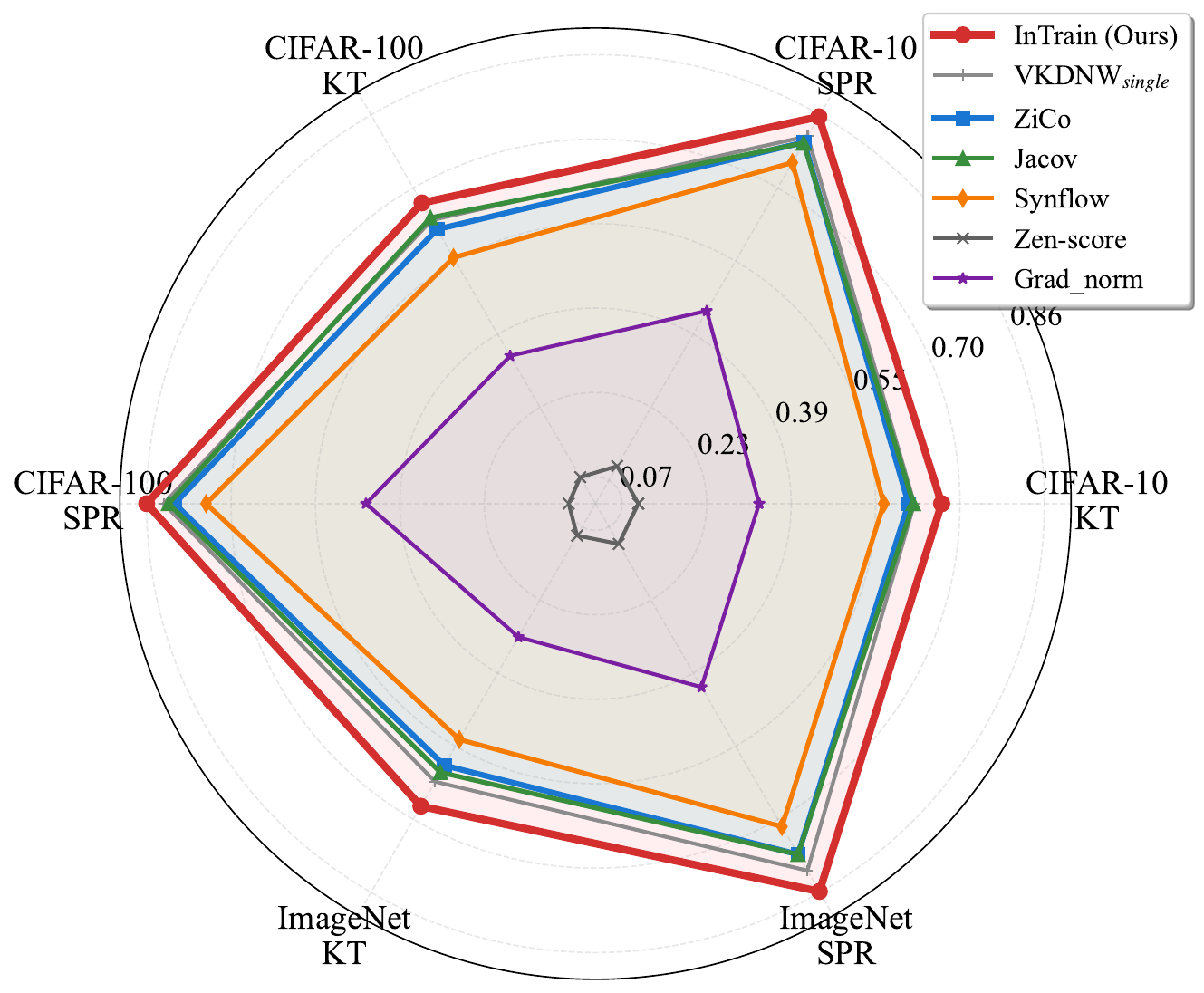}
\caption{Performance comparison of InTrain against representative zero-cost proxies on NAS-Bench-201. KT and SPR denote Kendall's $\tau$ and Spearman correlation values, respectively.}
\label{fig:teaser}
\end{figure}
\input{sec/1_intro}
\input{sec/2_related}

\input{sec/3_method}

\input{sec/4_experiment}

\input{sec/5_conclusion}

\input{sec/6_acknowledge}
{
    \small
    \bibliographystyle{ieeenat_fullname}
    \bibliography{main}
}


\end{document}

%% file: sec/0_abstract.tex
\begin{abstract}
Training-free neural architecture search promises efficient discovery of high-performance networks without costly training. However, existing zero-cost proxies rely on fragmented heuristics that fail to capture the fundamental question: what makes an architecture trainable? This paper introduces Intrinsic Trainability (InTrain), a unified theoretical proxy that formalizes trainability as an architectural invariant emerging from two synergistic components: geometric capacity and optimization resilience.
We operationalize intrinsic trainability through analysis of neural information processing. Geometric capacity is quantified via the participation ratio of activation covariance eigenspectrum, capturing the effective dimensionality of representation manifolds. Optimization resilience is measured through cumulative gradient health, assessing the robustness of backpropagation across network depth.
InTrain synthesizes these dimensions through a scale-invariant multiplicative coupling, which we hypothesize is essential for capturing their synergistic, non-additive relationship.
Extensive experiments on standard NAS benchmarks and search spaces demonstrate that InTrain achieves ranking correlations on par with state-of-the-art ensemble-based proxies and outperforms other single-metric methods.
\end{abstract}

%% file: sec/1_intro.tex
\section{Introduction}
\label{sec:intro}
The success of deep learning has driven intense interest in automating neural architecture design. Neural Architecture Search (NAS) has demonstrated remarkable capability in discovering novel architectures that rival or surpass human-designed networks~\cite{zoph2017neural, liu2018darts, tan2019efficientnet}. However, traditional NAS methods~\cite{zoph2017neural, real2019regularized} require training thousands of candidate architectures to convergence, incurring prohibitive computational costs—often thousands of GPU days. This computational barrier has motivated the development of training-free NAS proxies~\cite{mellor2021neural, abdelfattah2021zerocost, lin2021zen} that predict architecture quality without any optimization.

Training-free proxies~\cite{abdelfattah2021zerocost,tanaka2020pruning,lee2019snip,wang2020picking,shu2021nasi,ning2021evaluating,zhang2021gradsign} reduce evaluation costs by several orders of magnitude. Recent methods employ diverse zero-cost metrics based on activation statistics~\cite{mellor2021neural}, gradient properties~\cite{abdelfattah2021zerocost}, network expressivity~\cite{lin2021zen}, and synaptic diversity~\cite{tanaka2020pruning}. While promising, these approaches share a fundamental limitation that they rely on isolated heuristics without addressing what fundamentally makes an architecture trainable. Existing proxies examine symptoms, such as gradient magnitude or activation variance, rather than the underlying architectural properties governing trainability. This motivates a search for a unified, theoretically grounded criterion of trainability.

This fragmentation raises a fundamental question of whether a unified principle determines a network’s trainability. We argue the answer lies in formalizing intrinsic trainability, which is an architectural invariant independent of training procedures. Unlike empirical trainability, which depends on optimization algorithms and hyperparameters, intrinsic trainability is a fundamental property encoded in the topology and initial parametrization.

Intrinsic trainability emerges from two synergistic dimensions. First, an architecture should possess sufficient geometric capacity to represent rich, high-dimensional functions. We quantify this capacity using the Participation Ratio (PR)~\cite{stephenson2021geometry, martin2021implicit}, which measures the effective dimensionality of activation manifolds. Architectures with collapsed representations (low PR) cannot express the required complexity. Second, the architecture should exhibit optimization resilience for stable gradient propagation, addressing vanishing/exploding gradient problem~\cite{hochreiter1997long, pascanu2013difficulty}. This property, formalized by dynamical isometry~\cite{pennington2017resurrecting} and enabled by residual connections~\cite{he2016deep}, is essential for effective backpropagation. Crucially, we argue these properties are multiplicatively coupled: optimization resilience gates geometric capacity.
We introduce InTrain, a principled framework operationalizing this theory through rigorous neural information analysis. We quantify geometric capacity via the participation ratio of layer-wise activation covariance eigenspectra capturing information uniformity. Optimization resilience is assessed through cumulative gradient health, measuring backpropagation stability across network parameters.
We validate InTrain on NAS-Bench-101~\cite{ying2019bench}, NAS-Bench-201~\cite{dong2020nasbench201} and the MobileNetv2 search space. As shown in Fig.~\ref{fig:teaser}, InTrain achieves competitive ranking correlation
with ground-truth accuracy, outperforming most established single-metric proxies including ZiCo~\cite{li2023zico} and VKDNW$_{single}$~\cite{tybl2025training} on NAS-Bench-201. 
Our contributions are threefold:
\begin{itemize}
    \item We formalize intrinsic trainability as a unified architectural invariant, providing a principled theoretical framework for training-free architecture search.
    \item We propose InTrain, operationalizing intrinsic trainability through geometric capacity and optimization resilience, coupled via scale-invariant multiplicative interaction.
    \item We achieve competitive performance on NAS-Bench-101 and NAS-Bench-201, validating that our principled theory translates to empirical results on par with complex, ensemble-based methods.
\end{itemize}

%% file: sec/2_related.tex
\section{Related Work}
\label{sec:relate}
\noindent\textbf{Neural Architecture Search.}
Neural Architecture Search has fundamentally transformed how we design deep networks. Early methods employed reinforcement learning~\cite{zoph2017neural,zoph2018learning} or evolutionary algorithms~\cite{real2019regularized} to explore discrete search spaces, achieving remarkable results but requiring thousands of GPU days. Zoph \emph{et al.}~\cite{zoph2017neural} pioneered RL-based search, discovering architectures that matched hand-designed networks on ImageNet, while Real \emph{et al.}~\cite{real2019regularized} demonstrated that regularized evolution could discover state-of-the-art models. These methods established NAS as a viable paradigm but remained computationally prohibitive for most research settings.
Differentiable NAS~\cite{liu2018darts, xie2018snas} marked a paradigm shift by relaxing discrete search spaces into continuous ones, thereby enabling gradient-based optimization. DARTS~\cite{liu2018darts} reduced search costs from thousands to single-digit GPU days, sparking widespread adoption. One-shot methods~\cite{pham2018efficient, bender2018understanding, guo2020single,cai2018proxylessnas,cai2019once,hu2022generalizing} further improved efficiency through weight sharing: ENAS~\cite{pham2018efficient} trained a supernet once and sampled subnetworks for evaluation, while Single Path One-Shot~\cite{guo2020single} achieved uniform sampling for fair architecture comparison. However, weight-sharing introduces ranking inconsistencies~\cite{yu2020evaluating}, while differentiable formulations suffer from optimization instability~\cite{zela2020understanding}. These limitations, combined with the still-substantial training requirements, motivated the development of training-free proxies that can evaluate architectures in seconds without training.

\noindent\textbf{Training-Free NAS Proxies.}
Training-free (zero-cost) NAS proxies aim to predict network performance at initialization, avoiding the heavy cost of training. 
Early works focus on gradient-based proxies that exploit backpropagation dynamics.
SNIP~\cite{lee2019snip} and GraSP~\cite{wang2020picking} measure connection sensitivity through gradient saliency, while SynFlow~\cite{tanaka2020pruning} improves stability by propagating synthetic gradients via untrained networks.
Subsequent methods explore different facets of gradient flow: NASWOT~\cite{mellor2021neural} analyzes Neural Tangent Kernel (NTK) conditioning, while ZiCo~\cite{li2023zico} quantifies gradient statistics to model training dynamics.

In parallel, activation-based proxies analyze representational geometry through forward activations.
Zen-NAS~\cite{lin2021zen} evaluates Gaussian complexity and feature diversity, whereas TE-NAS~\cite{te-nas} employs trajectory entropy to approximate optimization difficulty.
More recently, ensemble methods such as AZ-NAS~\cite{lee2024az} combine multiple heterogeneous proxies to boost correlation, albeit at the cost of interpretability and computational simplicity.
Despite their empirical success, these proxies remain fragmented—gradient-based measures capture optimization sensitivity but ignore representational geometry, while activation-based metrics assess expressivity but neglect training dynamics.
Comprehensive benchmarking studies\cite{abdelfattah2021zerocost, tybl2025training} further show that many proxies exhibit inconsistent correlation across datasets and search spaces.
This fragmentation motivates the key question: what fundamental properties, unifying both geometry and gradients, determine whether an architecture can be effectively trained?

%% file: sec/3_method.tex
\section{Methodology}
\label{sec:method}
In this section, we introduce intrinsic trainability as the fundamental architectural property that quantifies a network's inherent capacity for effective optimization. 
We develop a principled framework grounded in information geometry and dynamical-systems theory, quantifying trainability through bidirectional information analysis.

\subsection{Theoretical Foundation}

Deep neural networks function as hierarchical information processors. During forward propagation, inputs are progressively transformed through a sequence of nonlinear mappings, each constructing increasingly abstract representations. During backpropagation, error signals flow backward through parameter space, enabling gradient-based optimization. A trainable architecture must excel in both directions: it must possess sufficient geometric capacity to represent complex functions, and sufficient optimization resilience to enable effective gradient flow.

We formalize intrinsic trainability through three design principles that any principled measure must satisfy. First, depth-invariance enables fair comparison across architectures of varying depths. A 50-layer network naturally accumulates more capacity than a 10-layer network through sheer depth. A meaningful comparison requires normalization by topological complexity. Second, compositionality reflects the hierarchical nature of deep learning. Information flows multiplicatively through layers: a bottleneck at any stage constrains all downstream processing. This suggests log-product rather than arithmetic aggregation. Third, bidirectionality recognizes that both forward and backward passes are essential. Following these principles, we separately quantify geometric capacity $\gamma(\mathcal{A})$ and optimization resilience $o(\mathcal{A})$, then couple them multiplicatively to obtain intrinsic trainability $I(\mathcal{A})$. Detailed formulations for each component are provided in the following subsections.

\subsection{Geometric Capacity via Participation Ratio}

The forward pass transforms inputs through a sequence of representation manifolds. At layer $\ell$, activations $\mathbf{A}_\ell\in\mathbb{R}^{N \times C}$ (for $N$ samples and $C$ channels) form a point cloud in the feature space $\mathbb{R}^C$. 
The effective dimensionality of this cloud determines the layer's representational power, which we analyze through the activation covariance structure. 
The covariance matrix $\mathbf{C}_\ell $ captures second-order statistics, and its eigenspectrum reveals how variance is distributed across dimensions. If all variance concentrates in a single eigenmode, the layer effectively collapses information, whereas a uniform spectrum indicates information preserved across many channels~\cite{ma2024outlier}. 
Standard measures like matrix rank are inadequate as rank treats all non-zero singular values equally, ignoring their magnitudes. Thus, we require a metric that captures this uniformity of information distribution.

\paragraph{Participation Ratio as Effective Dimensionality.}
Instead we use the participation ratio (PR)~\cite{stephenson2021geometry} as an effective dimensionality measure. For covariance matrix $\mathbf{C}_\ell$ with eigenvalues $\{\lambda_i\}_{i=1}^C$:
\begin{equation}
\text{PR}(\mathbf{C}_\ell)  = \frac{(\text{Tr}~\mathbf{C}_\ell)^2}{\text{Tr}(\mathbf{C}_\ell^2)} = \frac{(\sum_{i=1}^{C} \lambda_i)^2}{\sum_{i=1}^{C} \lambda_i^2}
\label{eq:pr}
\end{equation}

The participation ratio is related to the R\'enyi entropy of order 2 for the normalized eigenvalue distribution. Specifically, if $p_i = \lambda_i / \sum_j \lambda_j = \lambda_i / \operatorname{Tr}(\mathbf{C}_\ell)$ are the normalized eigenvalues, $\text{PR} = 1 / \sum_i p_i^2 = \exp(H_2)$ where $H_2 = -\log \sum_i p_i^2$ is the R\'enyi entropy. This provides a principled information-theoretic interpretation that high PR indicates variance spread across many dimensions~\cite{zheng2025information}. More analysis is provided in the supplement.

Given participation ratios $\{\text{PR}(\mathbf{C}_\ell)\}_{\ell=1}^L$ for all layers, how do we aggregate them into a single capacity score? The compositionality principle provides guidance. In deep networks, information flows hierarchically: layer $\ell$ transforms representations from layer $\ell-1$, and these transformed representations feed into layer $\ell+1$. Critically, this flow is multiplicative in its constraints. If layer $\ell$ has low PR (collapsed representation), it creates a bottleneck that limits all subsequent layers' representational power, regardless of their individual capacities.
This motivates us to adopt log-product aggregation. By defining the network's capacity as the product of layer-wise capacities ($\text{Capacity} \propto \prod_{\ell=1}^L \text{PR}(\mathbf{C}_\ell)$), and taking the logarithm, we obtain the total geometric capacity $\gamma(\mathcal{A})$ as a tractable sum:
\begin{equation}
\gamma(\mathcal{A}) = \sum_{\ell=1}^{L} \log (\text{PR}(\mathbf{C}_\ell))
\label{eq:gamma}
\end{equation}

During a single forward pass with batch $\mathbf{X} \in \mathbb{R}^{N \times D}$, we capture activation tensors $\mathbf{A}_\ell$ at each layer. For convolutional layers producing spatial activations of shape $(N, C, H, W)$, we reshape to $(N \cdot H \cdot W, C)$ to treat spatial locations as additional samples, computing covariance over the channel dimension. This is consistent with the interpretation that channels represent distinct features learned by the network.
For each layer, we compute the centered covariance matrix:
\begin{equation}
\mathbf{C}_\ell = \frac{1}{N_{\text{eff}}}(\mathbf{A}_\ell - \bar{\mathbf{A}}_\ell)^\top(\mathbf{A}_\ell - \bar{\mathbf{A}}_\ell) + \epsilon \mathbf{I}
\end{equation}
where $\bar{\mathbf{A}}_\ell$ is the mean activation, $N_{\text{eff}} = N \cdot H \cdot W$ for convolutional layers, $\mathbf{I}$ is the identity matrix, and $\epsilon = 10^{-10}$ provides numerical stabilization. The regularization term $\epsilon \mathbf{I}$ prevents singular matrices while being small enough not to affect the eigenspectrum of well-conditioned matrices.

\subsection{Optimization Resilience via Gradient Health}
Geometric capacity characterizes what the network can represent, while optimization resilience determines if it can be learned. An expressive network is untrainable if gradients exhibit pathological behaviors; we assess this resilience via gradient analysis.

\paragraph{Gradient Pathologies and Stability.}
The vanishing and exploding gradient problems~\cite{hochreiter1997long, pascanu2013difficulty,ma2023solving} have been fundamental challenges in deep learning for decades. When gradients vanish, parameters in early layers receive negligible updates, preventing the network from learning long-range dependencies. When gradients explode, training becomes unstable with divergent parameter updates. Both pathologies stem from the multiplicative nature of backpropagation: gradients at layer $\ell$ depend on the product of Jacobians from all downstream layers.
Architectures like ResNets~\cite{he2016deep} address these issues through careful design: residual connections provide gradient highways, batch normalization~\cite{ioffe2015batch} controls activation scales. However, in architecture search, we evaluate arbitrary topologies that may lack such safeguards. We require a metric that quantifies gradient health without assuming specific architectural motifs.

\paragraph{Variance-to-Maximum Ratio as Health Indicator.}
To quantify gradient health without relying on training dynamics, we define a layer-agnostic metric based on the statistical dispersion of gradients. For each parameter $\theta_i$, the gradient $\nabla_{\theta_i} \mathcal{L}$ indicates how the loss changes with respect to $\theta_i$. A healthy gradient should exhibit well-distributed magnitudes across its elements rather than concentration in a few components, and its overall scale should remain within a stable range. We capture these properties through a single metric:
\begin{equation}
h(\theta_i) = \min\left(1, \frac{\sigma(\nabla_{\theta_i})}{\max (|\nabla_{\theta_i}|) + \epsilon}\right)
\label{eq:health}
\end{equation}
where $\sigma(\cdot)$ denotes standard deviation over gradient elements and $\epsilon = 10^{-10}$ prevents division by zero. This variance-to-maximum ratio measures how uniformly gradient magnitude distributes across parameter elements. High ratios indicate balanced gradients where many elements contribute comparably, reflecting stable optimization, while low ratios suggest instability.
The capping operation is set to 1 to prevent outliers from dominating and maintains interpretability within a bounded scale. More analysis is provided in the supplement.

This formulation is conceptually related to the gradient conflict literature in multi-task learning~\cite{yu2020gradient}, where conflicting gradients indicate optimization challenges. Our variance-to-maximum ratio captures a related phenomenon: high variance relative to maximum indicates diversity in gradient directions across parameter elements, which generally correlates with stable, well-conditioned optimization.

Unlike geometric capacity, which compounds multiplicatively as a serial process (a chain limited by its weakest link), we posit that optimization resilience is an additive, parallel property. The network's total capacity to accept stable gradient updates is the sum of all available, independent parameter pathways, not a product constrained by the worst one. This motivates a cumulative summation rather than a log-product aggregation:
\begin{equation}
o(\mathcal{A}) = \sum_{i=1}^{|\Theta|} h(\theta_i)
\label{eq:rho}
\end{equation}
where $|\Theta|$ is the number of parameters. 
This cumulative measure reflects a key insight that optimization resilience represents the total "capacity" of the network to accept gradient updates. Each healthy parameter contributes a unit of this capacity; unhealthy parameters contribute less.

\paragraph{Synthetic Loss Design.}
Computing gradients requires a loss function. 
To isolate the effect of the synthetic gradient mechanism from data-distribution effects, we employ dummy inputs and randomly generated targets. 
Specifically, we generate synthetic input batches $\mathbf{X} \sim \mathcal{N}(0, 1)$ and synthetic target labels $\mathbf{t} \sim \mathcal{U}(0, C)$, where $C$ denotes the number of output classes. 
This setup ensures that gradient signals are architecture-dependent rather than dataset-dependent, allowing us to probe intrinsic gradient dynamics without real data supervision.
We then use a simple synthetic loss that activates all network paths without architecture-specific design. For network output $\mathbf{y} \in \mathbb{R}^{N \times K}$ (batch size $N$, output dimension $K$), we generate random targets and compute loss. If $K > 1$, indicating classification, we sample random class labels $\mathbf{t} \sim \text{Categorical}(1/K)$ and use cross-entropy: 
\begin{equation}
\mathcal{L}_C = -\sum_{n,k} \mathbb{I}[t_n = k] \log y_{nk},
\label{eq:lss}
\end{equation}

If the output is spatial (\emph{e.g.}, segmentation with shape $(N, K, H', W')$), we generate a random tensor $\mathbf{t}$ of matching shape and use mean-square error: 
\begin{equation}
\mathcal{L}_M = \|\mathbf{y} - \mathbf{t}\|^2,
\label{eq:lossmse}
\end{equation}

This synthetic loss design ensures gradients flow through all parameters regardless of task or architecture topology. Critically, we do not require real labels or task-specific loss functions—the goal is not to measure performance but to probe gradient dynamics. 
The random targets provide a generic optimization signal that exercises all pathways through the network, revealing architectural properties that affect gradient propagation.

\subsection{Intrinsic Trainability}

Having separately quantified geometric capacity $\gamma(\mathcal{A})$ and optimization resilience $o(\mathcal{A})$, we now address a central question: how do these properties jointly determine a network’s intrinsic trainability?

\paragraph{The Multiplicative Gating Hypothesis.}
We hypothesize that a network’s capacity and resilience are not independent factors but interact multiplicatively through a gating mechanism. Consider two limiting cases. When geometric capacity $\gamma$ is large but optimization resilience $o \approx 0$, the network possesses rich representational power but cannot propagate gradients—learning collapses despite high expressivity. Conversely, when $o$ is large but $\gamma$ approaches a degenerate limit, optimization remains stable, but the model can only represent trivial mappings.
These cases reveal an intrinsic asymmetry: capacity without resilience cannot be realized, and resilience without capacity achieves nothing meaningful. 
In other words, each component functions as a gate for the other.
Consequently, a multiplicative interaction ($\gamma \times (1 + o)$) better captures trainability than an additive formulation ($\gamma + o$), which would incorrectly assign moderate trainability even when one factor collapses.

\paragraph{Formulation and Normalization.}
We define the intrinsic trainability of an architecture $\mathcal{A}$ with depth $L$ as:
\begin{equation}
I(\mathcal{A}) = \frac{\gamma(\mathcal{A}) \times (1 + o(\mathcal{A}))}{\log(L + 1)},
\label{eq:tau}
\end{equation}
The term $(1 + o)$ guarantees positivity and prevents degenerate cases when $o = 0$, while preserving proportional scaling with resilience. The logarithmic normalization by $\log(L+1)$ ensures depth-invariant comparison across architectures. Since both $\gamma$ and $o$ tend to increase with depth, direct normalization by $L$ would overcompensate, suppressing depth’s legitimate contribution from deeper networks. The logarithmic form instead reflects the sublinear growth of information processing capacity with depth and empirically stabilizes cross-depth comparisons.

%% file: sec/4_experiment.tex
\section{Experiments}
\label{sec:experiments}
In this section, we conduct a comprehensive empirical evaluation assessing InTrain's ranking fidelity and utility in complete NAS pipelines. We begin by introducing the benchmarks and search spaces. We then present the implementation details of our method. Next, we compare the proposed method with state-of-the-art methods on NASBench-101 \cite{ying2019bench}, NASBench-201 \cite{dong2020nasbench201}, and MobileNetV2~\cite{sandler2018mobilenetv2,lin2021zen} search spaces.
Finally, we perform critical ablation studies to dissect our method and verify the synergistic contribution of its constituent components.

\begin{figure*}[t]
    \centering
    \includegraphics[width=0.9\linewidth]{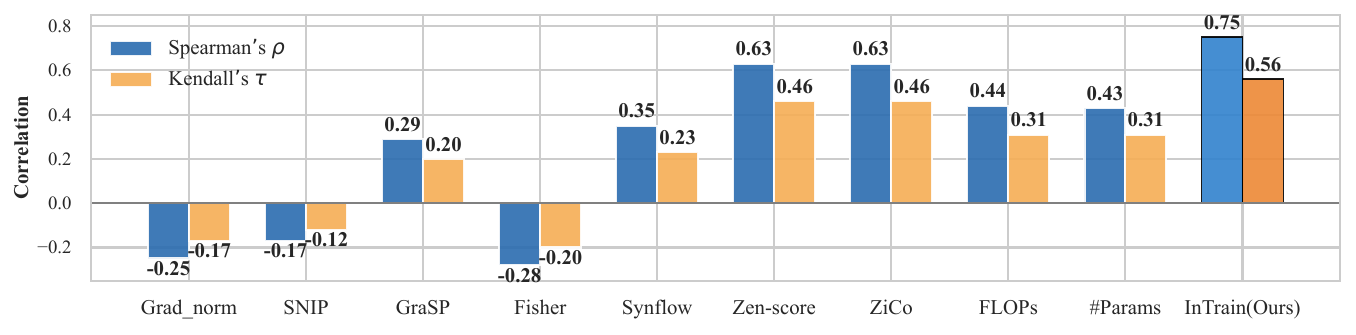}
    \caption{Comparison of correlation coefficients between zero-cost proxies and final test accuracy on NASBench-101.}
    \label{fig:correlation_comparison}
\end{figure*}

\begin{table*}[!t]
    \centering
    \caption{Comparison of our InTrain against different zero-cost proxies on NASBench-201. KT and SPR denote Kendall's $\tau$ and Spearman correlation values, respectively.}
    \label{tab:3-1}
    \vspace{5mm}
    \setlength{\tabcolsep}{7.3mm}
    \begin{tabular}{lcccccc}
    \toprule
    \multirow{2}{*}{Proxies} & \multicolumn{2}{c}{CIFAR-$10$} & \multicolumn{2}{c}{CIFAR-$100$} & \multicolumn{2}{c}{ImageNet16-120} \\ 
    \cmidrule(lr){2-3} \cmidrule(lr){4-5} \cmidrule(lr){6-7}
                            & KT & SPR & KT & SPR & KT & SPR \\ 
    \midrule
    Grad\_norm~\cite{abdelfattah2021zerocost} & 0.328 & 0.438 & 0.341 & 0.451 & 0.310 & 0.418 \\ 
    SNIP~\cite{lee2019snip}       & 0.431 & 0.591 & 0.440 & 0.597 & 0.389 & 0.521 \\ 
    GraSp~\cite{wang2020picking}      & 0.352 & 0.505 & 0.349 & 0.498 & 0.359 & 0.502 \\ 
    Fisher~\cite{liu2021group}     & 0.400 & 0.550 & 0.410 & 0.550 & 0.370 & 0.500 \\ 
    Synflow~\cite{tanaka2020pruning}    & 0.561 & 0.758 & 0.553 & 0.750 & 0.531 & 0.719 \\ 
    TE-NAS~\cite{te-nas}         & 0.536 & 0.722 & 0.537 & 0.723 & 0.523 & 0.709 \\ 
    Jacov~\cite{abdelfattah2021zerocost}          & 0.616 & 0.800 & 0.639 & 0.820 & 0.602 & 0.779 \\ 
    NASWOT~\cite{mellor2021neural}         & 0.571 & 0.762 & 0.607 & 0.799 & 0.605 & 0.794 \\ 
    Zen-score~\cite{lin2021zen}  & 0.102 & 0.103 & 0.079 & 0.072 & 0.091 & 0.109 \\ 
    ZiCo~\cite{li2023zico}       & 0.607 & 0.802 & 0.614 & 0.809 & 0.587 & 0.779 \\ 
    AZ-NAS~\cite{lee2024az}         & 0.712 & 0.892 & 0.696 & 0.880 & 0.673 & 0.859 \\ 
    VKDNW$_{single}$~\cite{tybl2025training}          & 0.618 & 0.815 & 0.634 & 0.829 & 0.622 & 0.814 \\ 
    FLOPs               & 0.623 & 0.799 & 0.586 & 0.763 & 0.545 & 0.718 \\  
    \midrule
    \textbf{InTrain (ours)} & \textbf{0.669} & \textbf{0.857} & \textbf{0.671} & \textbf{0.861} & \textbf{0.675} & \textbf{0.859} \\
    \bottomrule
    \end{tabular}
\end{table*}

\subsection{Benchmarks and search spaces}

\begin{itemize}
    \item \textbf{NASBench-101~\cite{ying2019bench}:} A public benchmark containing 423,000 convolutional architectures represented as directed acyclic graphs with up to 7 nodes and 9 edges. Each cell employs three operations (3×3 conv, 1×1 conv, 3×3 max-pool) and architectures are evaluated on CIFAR-10 with comprehensive performance metrics.
    
    \item \textbf{NASBench-201~\cite{dong2020nasbench201}:} Comprises 15,625 architectures with 4-node cells featuring five operations (zero, skip connection, 3×3 conv, 1×1 conv, 3×3 avg-pool). Each architecture is evaluated on CIFAR-10, CIFAR-100, and ImageNet16-120, providing detailed accuracy and efficiency metrics.
    
    \item \textbf{MobileNetV2~\cite{sandler2018mobilenetv2,lin2021zen}:} An open-domain search space based on MobileNetV2's inverted residual blocks, some incorporating squeeze-and-excitation modules. The search space allows optimization of expansion ratios, kernel sizes, strides, layer counts, channel dimensions, and input resolutions.
\end{itemize}

\subsection{Implementation details} 
For InTrain, we use a batch of 64 synthetic images of resolution 64$\times$64, where each pixel is independently sampled from a standard Gaussian distribution. Randomly generated target labels are used during backward computation.
We primarily report Spearman's rank correlation coefficient ($\rho$) and Kendall's Tau ($\tau$) between proxy ranking and final validation accuracy after full training, consistent with prior works. 
To demonstrate the practical utility of InTrain in a complete NAS pipeline, we integrate it into an evolutionary search framework, referred to as InTrain-NAS. The hyperparameters and configuration of InTrain-NAS are based on common practices in evolutionary NAS literature~\cite{li2023zico,lee2024az,tybl2025training} and surveys on evolutionary neural architecture search~\cite{liu2021survey}.
More training details are provided in the supplement.

\begin{table*}[t]
\centering
\caption{Comparison of the searched results of our InTrain-based NAS against SOTA NAS methods on ImageNet under various FLOP budgets. For the `Method' column, `MS' means multi-shot NAS; `OS' is short for one-shot NAS; Scaling represents network scaling methods; `ZS' is short for zero-shot NAS. $^+$ denotes training from scratch.}
\label{tab:imagenet_comparison}
\setlength{\tabcolsep}{5.8mm}
\begin{tabular}{lccccc}
\toprule
\#FLOPs & Approach & FLOPs & Top-1 (\%) & Method & Costs [GPU Days] \\
\midrule
\multirow{9}{*}{ 450M} 
 & EfficientNet-B0~\cite{tan2019efficientnet}       & 390M & 77.1 & Scaling & 3800 \\
 & OFA$^+$~\cite{cai2019once}               & 406M & 77.7 & OS & 50 \\
 & DONNA~\cite{moons2021distilling}                 & 501M & 78.0 & OS & 25 \\
 & MnasNet-A3~\cite{tan2019mnasnet}            & 403M & 76.7 & MS & -- \\
 & BN-NAS~\cite{chen2021bn}                & 470M & 75.7 & MS & 0.8 \\
 & NASNet-B~\cite{zoph2018learning}              & 488M & 72.8 & MS & 1800 \\
 & CARS-D~\cite{yang2020cars}                & 496M & 73.3 & MS & 0.4 \\
 & ZiCo~\cite{li2023zico}                  & 448M & 78.1 & ZS & 0.4 \\
 & AZ-NAS~\cite{lee2024az}                   & 462M & 78.6 & ZS & 0.4 \\
 & VKDNW$_{agg}$~\cite{tybl2025training}          & 480M & 78.8 & ZS & 0.4 \\
 & \textbf{InTrain-NAS (Ours)}       & \textbf{455M} & \textbf{78.9} & \textbf{ZS} & \textbf{0.4} \\
\cmidrule{1-6}
\multirow{11}{*}{600M}
 & EfficientNet-B1~\cite{tan2019efficientnet}       & 700M & 79.1 & Scaling & 3800 \\
 & DARTS~\cite{liu2018darts}                 & 574M & 73.3 & OS & 4 \\
 & PC-DARTS~\cite{xu2019pc}              & 586M & 75.8 & OS & 3.8 \\
 & BigNAS-L~\cite{yu2020bignas}              & 586M & 79.5 & OS & 2304$^\dagger$ \\
 & EnTranNAS~\cite{yang2021towards}             & 594M & 76.2 & OS & 2.1 \\
 & MAGIC-AT~\cite{xu2022analyzing}              & 598M & 76.8 & OS & 2 \\
 & DONNA~\cite{moons2021distilling}                 & 599M & 78.4 & OS & 25 \\
 & CARS-I~\cite{yang2020cars}                & 591M & 75.2 & MS & 0.4 \\
 & SemiNAS~\cite{luo2020semi}               & 599M & 76.5 & MS & 4 \\

 & Zen-score~\cite{lin2021zen}             & 611M & 79.1 & ZS & 0.5 \\
 & ZiCo~\cite{li2023zico}                  & 603M & 79.4 & ZS & 0.4\\
 & AZ-NAS~\cite{lee2024az}                   & 615M & 79.9 & ZS & 0.6 \\
  & \textbf{InTrain-NAS (Ours)}       & \textbf{607M} & \textbf{79.9} & \textbf{ZS} & \textbf{0.4} \\
\cmidrule{1-6}
\multirow{4}{*}{1000M}
 & EfficientNet-B2~\cite{tan2019efficientnet}       & 1000M & 80.1 & Scaling & 3800 \\
 & sharpDARTS~\cite{hundt2019sharpdarts}            & 950M & 76.0 & OS & -- \\
 & Zen-score~\cite{lin2021zen}             & 934M & 80.8 & ZS & 0.5 \\
 & ZiCo~\cite{li2023zico}                  & 1005M & 80.5 & ZS & 0.4 \\
 & AZ-NAS~\cite{lee2024az}                   & 1022M & 81.1 & ZS & 0.7 \\
 & \textbf{InTrain-NAS (Ours)}       & \textbf{1013M} & \textbf{81.3} & \textbf{ZS} & \textbf{0.4} \\
\bottomrule
\end{tabular}
\end{table*}

\subsection{Main Results on NAS-Bench-101}
In this section, we evaluate the correlation between various zero-cost proxies and the ground-truth test accuracy on NAS-Bench-101~\cite{ying2019bench}. As illustrated in Fig.~\ref{fig:correlation_comparison}, InTrain achieves competitive rank correlation with the ground-truth performance compared to existing zero-cost proxies. Specifically, InTrain achieves Kendall's $\tau$ of 0.56 and Spearman's $\rho$ of 0.75 on the full NAS-Bench-101 search space, outperforming several established zero-cost proxies and indicating strong predictive consistency.
The correlation results reveal that InTrain surpasses simple architectural metrics such as FLOPs (Kendall’s $\tau$ = 0.44) and parameter count (Kendall’s $\tau$ = 0.43). 
This improvement suggests that combining gradient dynamics with intermediate feature representations yields a more comprehensive evaluation framework.
It provides richer insight than approaches relying solely on structural attributes such as parameter counts or computational complexity.
Furthermore, InTrain shows competitive performance against state-of-the-art zero-cost proxies including ZiCo~\cite{li2023zico} (Kendall's $\tau$ = 0.46) and Zen-score~\cite{lin2021zen} (Kendall's $\tau$ = 0.31). 
Whereas prior proxies rely on either gradient statistics or activation-based heuristics in isolation, InTrain provides a holistic measure of architectural trainability by jointly modeling geometric capacity and optimization resilience.
This integrated approach appears to capture fundamental aspects of network behavior that contribute to final performance, providing a more reliable basis for architecture selection without requiring extensive training.
The consistent correlation patterns across different evaluation metrics (both Kendall's $\tau$ and Spearman's $\rho$) provide additional evidence of InTrain's robustness as a reliable zero-cost proxy for NAS. 
It achieves a balanced trade-off between predictive accuracy and computational efficiency.

\subsection{Main Results on NAS-Bench-201}
We provide a comprehensive correlation analysis on the NAS-Bench-201 benchmark~\cite{dong2020nasbench201}, which reveals the comparative effectiveness of various zero-cost proxies across multiple datasets. As detailed in Table~\ref{tab:3-1}, our evaluation encompasses CIFAR-10, CIFAR-100, and ImageNet16-120 to rigorously assess cross-dataset generalization capabilities.
The proposed InTrain method demonstrates superior ranking consistency, achieving Kendall's $\tau$ of 0.669, 0.671, and 0.675 on CIFAR-10, CIFAR-100, and ImageNet16-120, respectively. Corresponding Spearman correlations of 0.857, 0.861, and 0.859 further substantiate its robust predictive performance. These results indicate that InTrain effectively captures architectural quality while maintaining remarkable stability across diverse data distributions.

Among all evaluated proxies, 
the ensemble-based AZ-NAS~\cite{lee2024az} achieves the highest correlation scores on CIFAR-10 and CIFAR-100. 
This performance gain stems from its strategy of ensembling
multiple, disparate proxy metrics (\emph{e.g.}, expressivity, trainability, and complexity, as reported in~\cite{lee2024az}). However, this approach essentially represents an empirical aggregation of heterogeneous heuristics, lacking a unified theoretical foundation.
In contrast, InTrain achieves competitive performance using a single theoretically grounded proxy, outperforming all other non-ensemble methods. 
This confirms that our principled framework captures a more coherent and discriminative signal than fragmented heuristic combinations. Notably, ZiCo~\cite{li2023zico} also exhibits strong Spearman correlations, while Synflow performs particularly well on CIFAR-100. Traditional complexity-based metrics, such as FLOPs, remain competitive in terms of Spearman correlation, though they lack theoretical interpretability. 
We also observe that most proxies show noticeable performance fluctuations across different datasets, likely caused by their sensitivity to data distribution and sample composition. In contrast, InTrain maintains stable rankings across all datasets. This consistency likely stems from the theoretical grounding in intrinsic architectural properties, rather than reliance on dataset-dependent heuristics.

\subsection{Main Results on MobileNetV2}
To ensure fair comparison with established zero-shot NAS methods~\cite{li2023zico,lin2021zen,lee2024az,tybl2025training}, we adopt the standardized MobileNetV2 search space~\cite{sandler2018mobilenetv2,lin2021zen}. Based on InTrain, we conduct architecture search using an evolutionary algorithm to form InTrain-NAS. This setting allows us to evaluate the generalizability and practical applicability of InTrain on large-scale datasets and complex search spaces. All experiments are conducted on the ImageNet-1K benchmark~\cite{deng2009imagenet} under computational budgets constrained to 450M, 600M, and 1000M FLOPs.
As summarized in Table~\ref{tab:imagenet_comparison}, we report comprehensive results on ImageNet across diverse NAS paradigms under different computational budgets. The comparative analysis encompasses conventional scaling methods, multi-shot NAS (MS), one-shot NAS (OS), and emerging zero-shot (ZS) approaches.
InTrain-NAS consistently achieves superior performance across all computational regimes, achieving top-1 accuracies of 78.9\%, 79.9\%, and 81.3\% under 450M, 600M, and 1000M FLOP constraints, respectively. Notably, the architectures discovered by InTrain-NAS maintain competitive model complexity, with actual FLOPs closely matching the prescribed budgets. Despite this parity, they deliver consistently higher recognition accuracy.
These results demonstrate that InTrain-NAS achieves an effective balance between architectural expressivity and computational efficiency, validating its capacity to discover high-performance architectures with minimal computational cost.

\begin{table}[t]
\centering
\caption{Component ablation (NAS-Bench-201).}
\label{tab:ablation_components}
\setlength{\tabcolsep}{2.5mm}
\begin{tabular}{lcc}
\toprule
Variant & Kendall's $\tau$ & Spearman $\rho$ \\
\midrule
PR-only & 0.61 & 0.82 \\
Grad-only & 0.63 & 0.83\\
PR + Grad (additive) & 0.59 & 0.80 \\
\textbf{InTrain (ours)} & \textbf{0.67} & \textbf{0.86} \\
\bottomrule
\end{tabular}
\end{table}

\subsection{Ablation Studies}
We perform extensive ablation studies to disentangle the contribution of individual components and investigate how different input regimes affect the robustness of InTrain.

\paragraph{Component ablation.} 
To validate the design of InTrain, we conduct a comprehensive component ablation study as shown in Table~\ref{tab:ablation_components}. The results show that both geometric capacity (PR-only) and gradient health (Grad-only) individually provide meaningful information, achieving Kendall's $\tau$ of 0.61 and 0.63, respectively. A naive additive combination (PR + Grad) results in a Kendall's $\tau$ of 0.59, which is substantially worse than either component alone. This finding empirically supports our core hypothesis that geometric capacity and optimization resilience are not independent additive factors. Their naive summation introduces interference between the two signals, whereas our proposed multiplicative coupling (InTrain, $\tau=0.67$) effectively captures their synergistic relationship.
InTrain achieves consistent improvement, with Kendall's $\tau$ of 0.67 and Spearman $\rho$ of 0.86, showing clear gains over both individual components and the additive baseline. This performance gap highlights the importance of our carefully designed integration strategy, which enables synergistic interaction between geometric capacity and gradient health metrics.

\paragraph{Robustness to Input Variations.}
We evaluate InTrain under diverse synthetic-input regimes on NAS-Bench-201 with CIFAR-10 as the validation set. Specifically, we consider Gaussian inputs, all-ones, uniform inputs, and varying input resolutions (32$\times$32, 64$\times$64, 128$\times$128). Unless otherwise noted, the default protocol uses batch size $B=64$ and Gaussian inputs at resolution $R=64$.

\begin{figure}[t]
    \centering
    \includegraphics[width=0.9\linewidth]{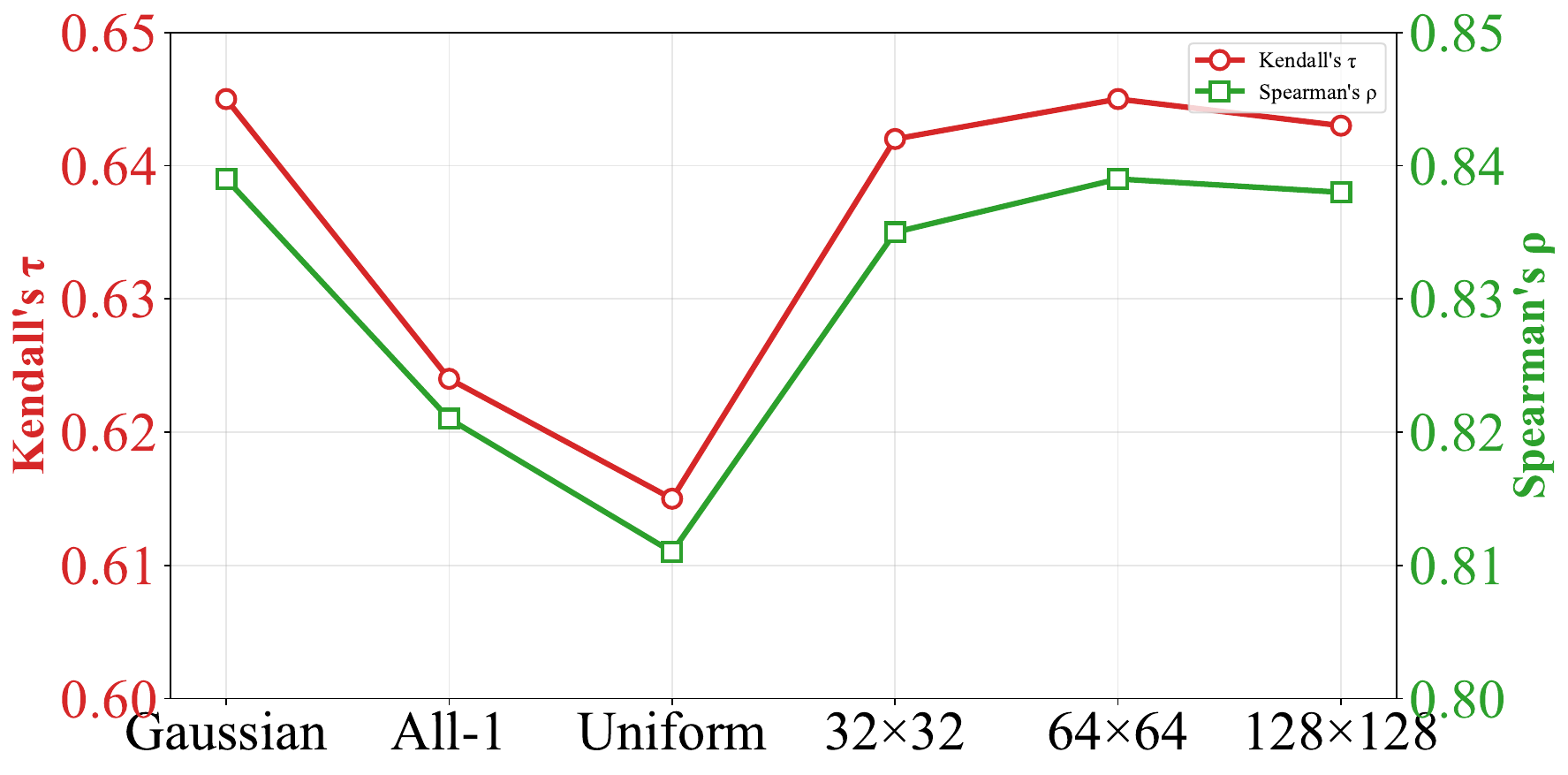}
    \caption{Stability analysis of InTrain under different input data.}
    \label{fig:stab}
\end{figure}

As illustrated in Figure~\ref{fig:stab}, InTrain maintains high and stable correlation across tested resolutions under our default protocol, indicating robustness to input scale in practical settings. 
We observe that Gaussian inputs tend to produce more stable covariance estimates and slightly higher correlation than uniform noise. A plausible explanation is that Gaussian perturbations create smoother, lower-magnitude activation variations that facilitate consistent eigenstructure estimation. In contrast, uniform noise often induces large outliers, resulting in noisier covariances.
Constant all-one inputs without perturbation lead to degenerate, near-singular covariance matrices, and are therefore not recommended.
These findings confirm InTrain’s robustness across varying input conditions. Moreover, they also demonstrate that simple synthetic inputs suffice for efficient architecture evaluation without sacrificing correlation quality.

%% file: sec/5_conclusion.tex
\section{Conclusion}
\label{sec:conclusion}
We introduce Intrinsic Trainability (InTrain), a principled zero-cost proxy that formalizes trainability as a multiplicative coupling between geometric capacity and optimization resilience. Geometric capacity is measured via the participation ratio of layer-wise activation covariances, while optimization resilience is captured by a variance-to-maximum gradient-health metric. InTrain combines these two components to efficiently evaluate architectures using a single forward and backward pass on a synthetic minibatch. Empirically, InTrain yields competitive ranking correlations on NAS-Bench datasets and SOTA searched models under FLOP budgets while requiring only modest compute (0.4 GPU days for InTrain-NAS in our setup). 
Compared with prior zero-cost proxies, InTrain provides both high predictive fidelity and layer-wise diagnostics on capacity limits and gradient issues, improving interpretability for architecture selection.

%% file: sec/6_acknowledge.tex
\noindent\textbf{Acknowledgments:}

This work was supported by the Research Initiation Fund Project of Fuzhou University (No.511611), the Fujian Provincial Department of Education Youth Project (No.JZ230006), the National Natural Science Foundation of China (No.62306306, No.62506146), GuangDong Basic and Applied Basic Research Foundation (No.2022A1515110128), Jiangxi Provincial Natural Science Foundation (No.20252BAC200196), Early-Career Young Scientists and Technologists Project of Jiangxi Province (No.20252BEJ730022).

%% file: main.bib
@String(AAAI = {AAAI})

@inproceedings{zoph2017neural,
  title={Neural architecture search with reinforcement learning},
  author={Zoph, Barret and Le, Quoc V},
  booktitle={International Conference on Learning Representations},
  year={2017}
}

@inproceedings{liu2018darts,
  title={DARTS: Differentiable architecture search},
  author={Liu, Hanxiao and Simonyan, Karen and Yang, Yiming},
  booktitle={International Conference on Learning Representations},
  year={2019}
}

@inproceedings{tan2019efficientnet,
  title={Efficientnet: Rethinking model scaling for convolutional neural networks},
  author={Tan, Mingxing and Le, Quoc},
  booktitle={International conference on machine learning},
  pages={6105--6114},
  year={2019},
  organization={PMLR}
}

@inproceedings{real2019regularized,
  title={Regularized evolution for image classifier architecture search},
  author={Real, Esteban and Aggarwal, Alok and Huang, Yanping and Le, Quoc V},
  booktitle={Proceedings of the aaai conference on artificial intelligence},
  volume={33},
  number={01},
  pages={4780--4789},
  year={2019}
}

@inproceedings{mellor2021neural,
  title={Neural architecture search without training},
  author={Mellor, Joe and Turner, Jack and Storkey, Amos and Crowley, Elliot J},
  booktitle={International conference on machine learning},
  pages={7588--7598},
  year={2021},
  organization={PMLR}
}

@inproceedings{abdelfattah2021zerocost,
  title={Zero-cost proxies for lightweight NAS},
  author={Abdelfattah, Mohamed S and Mehrotra, Abhinav and Dudziak, {\L}ukasz and Lane, Nicholas D},
  booktitle={International Conference on Learning Representations},
  year={2021}
}

@inproceedings{lin2021zen,
  title={Zen-nas: A zero-shot nas for high-performance image recognition},
  author={Lin, Ming and Wang, Pichao and Sun, Zhenhong and Chen, Hesen and Sun, Xiuyu and Qian, Qi and Li, Hao and Jin, Rong},
  booktitle={Proceedings of the IEEE/CVF international conference on computer vision},
  pages={347--356},
  year={2021}
}

@article{tanaka2020pruning,
  title={Pruning neural networks without any data by iteratively conserving synaptic flow},
  author={Tanaka, Hidenori and Kunin, Daniel and Yamins, Daniel L and Ganguli, Surya},
  journal={Advances in neural information processing systems},
  volume={33},
  pages={6377--6389},
  year={2020}
}

@inproceedings{ying2019bench,
  title={Nas-bench-101: Towards reproducible neural architecture search},
  author={Ying, Chris and Klein, Aaron and Christiansen, Eric and Real, Esteban and Murphy, Kevin and Hutter, Frank},
  booktitle={International conference on machine learning},
  pages={7105--7114},
  year={2019},
  organization={PMLR}
}

@inproceedings{dong2020nasbench201,
  title={NAS-Bench-201: Extending the scope of reproducible neural architecture search},
  author={Dong, Xuanyi and Yang, Yi},
  booktitle={International Conference on Learning Representations},
  year={2020}
}

@inproceedings{stephenson2021geometry,
  title={On the geometry of generalization and memorization in deep neural networks},
  author={Stephenson, Cory and Padhy, Suchismita and Ganesh, Abhinav and Hui, Yue and Tang, Hanlin and Chung, SueYeon},
  booktitle={International Conference on Learning Representations},
  year={2021}
}

@article{martin2021implicit,
  title={Implicit self-regularization in deep neural networks: Evidence from random matrix theory and implications for learning},
  author={Martin, Charles H and Mahoney, Michael W},
  journal={Journal of Machine Learning Research},
  volume={22},
  number={165},
  pages={1--73},
  year={2021}
}

@inproceedings{xie2018snas,
  title={SNAS: Stochastic neural architecture search},
  author={Xie, Sirui and Zheng, Hehui and Liu, Chunxiao and Lin, Liang},
  booktitle={International Conference on Learning Representations},
  year={2019}
}

@inproceedings{pham2018efficient,
  title={Efficient neural architecture search via parameters sharing},
  author={Pham, Hieu and Guan, Melody and Zoph, Barret and Le, Quoc and Dean, Jeff},
  booktitle={International conference on machine learning},
  pages={4095--4104},
  year={2018},
  organization={PMLR}
}

@inproceedings{bender2018understanding,
  title={Understanding and simplifying one-shot architecture search},
  author={Bender, Gabriel and Kindermans, Pieter-Jan and Zoph, Barret and Vasudevan, Vijay and Le, Quoc},
  booktitle={International conference on machine learning},
  pages={550--559},
  year={2018},
  organization={PMLR}
}

@inproceedings{guo2020single,
  title={Single path one-shot neural architecture search with uniform sampling},
  author={Guo, Zichao and Zhang, Xiangyu and Mu, Haoyuan and Heng, Wen and Liu, Zechun and Wei, Yichen and Sun, Jian},
  booktitle={European conference on computer vision},
  pages={544--560},
  year={2020},
  organization={Springer}
}

@inproceedings{yu2020evaluating,
  title={Evaluating the search phase of neural architecture search},
  author={Yu, Kaicheng and Sciuto, Christian and Jaggi, Martin and Musat, Claudiu and Salzmann, Mathieu},
  booktitle={International Conference on Learning Representations},
  year={2020}
}

@inproceedings{zela2020understanding,
  title={Understanding and robustifying differentiable architecture search},
  author={Zela, Arber and Elsken, Thomas and Saikia, Tonmoy and Marrakchi, Yassine and Brox, Thomas and Hutter, Frank},
  booktitle={International Conference on Learning Representations},
  year={2020}
}

@inproceedings{li2023zico,
  title={Zico: Zero-shot nas via inverse coefficient of variation on gradients},
  author={Li, Guihong and Yang, Yuedong and Bhardwaj, Kartikeya and Marculescu, Radu},
  journal={International Conference on Learning Representations},
  year={2023}
}

@article{hochreiter1997long,
  title={Long short-term memory},
  author={Hochreiter, Sepp and Schmidhuber, J{\"u}rgen},
  journal={Neural computation},
  volume={9},
  number={8},
  pages={1735--1780},
  year={1997},
  publisher={MIT press}
}

@inproceedings{pascanu2013difficulty,
  title={On the difficulty of training recurrent neural networks},
  author={Pascanu, Razvan and Mikolov, Tomas and Bengio, Yoshua},
  booktitle={International conference on machine learning},
  pages={1310--1318},
  year={2013},
  organization={Pmlr}
}

@inproceedings{he2016deep,
  title={Deep residual learning for image recognition},
  author={He, Kaiming and Zhang, Xiangyu and Ren, Shaoqing and Sun, Jian},
  booktitle={Proceedings of the IEEE conference on computer vision and pattern recognition},
  pages={770--778},
  year={2016}
}

@article{pennington2017resurrecting,
  title={Resurrecting the sigmoid in deep learning through dynamical isometry: theory and practice},
  author={Pennington, Jeffrey and Schoenholz, Samuel and Ganguli, Surya},
  journal={Advances in neural information processing systems},
  volume={30},
  year={2017}
}

@inproceedings{ioffe2015batch,
  title={Batch normalization: Accelerating deep network training by reducing internal covariate shift},
  author={Ioffe, Sergey and Szegedy, Christian},
  booktitle={International conference on machine learning},
  pages={448--456},
  year={2015},
  organization={pmlr}
}

@article{yu2020gradient,
  title={Gradient surgery for multi-task learning},
  author={Yu, Tianhe and Kumar, Saurabh and Gupta, Abhishek and Levine, Sergey and Hausman, Karol and Finn, Chelsea},
  journal={Advances in neural information processing systems},
  volume={33},
  pages={5824--5836},
  year={2020}
}

@inproceedings{lee2019snip,
  title={Snip: Single-shot network pruning based on connection sensitivity},
  author={Lee, Namhoon and Ajanthan, Thalaiyasingam and Torr, Philip HS},
  booktitle={International Conference on Learning Representations},
  year={2019}
}

@inproceedings{wang2020picking,
  title={Picking winning tickets before training by preserving gradient flow},
  author={Wang, Chaoqi and Zhang, Guodong and Grosse, Roger},
  booktitle={International Conference on Learning Representations},
  year={2020}
}

@inproceedings{zhang2021gradsign,
  title={Gradsign: Model performance inference with theoretical insights},
  author={Zhang, Zhihao and Jia, Zhihao},
  booktitle={International Conference on Learning Representations},
  year={2022}
}

@inproceedings{liu2021group,
  title={Group fisher pruning for practical network compression},
  author={Liu, Liyang and Zhang, Shilong and Kuang, Zhanghui and Zhou, Aojun and Xue, Jing-Hao and Wang, Xinjiang and Chen, Yimin and Yang, Wenming and Liao, Qingmin and Zhang, Wayne},
  booktitle={International Conference on Machine Learning},
  pages={7021--7032},
  year={2021},
  organization={PMLR}
}

@inproceedings{te-nas,
  title={Neural architecture search on imagenet in four gpu hours: A theoretically inspired perspective},
  author={Chen, Wuyang and Gong, Xinyu and Wang, Zhangyang},
  booktitle={International Conference on Learning Representations},
  year={2021}
}

@inproceedings{tybl2025training,
  title={Training-free neural architecture search through variance of knowledge of deep network weights},
  author={Tybl, Ondrej and Neumann, Lukas},
  booktitle={Proceedings of the Computer Vision and Pattern Recognition Conference},
  pages={14881--14890},
  year={2025}
}

@inproceedings{lee2024az,
  title={Az-nas: Assembling zero-cost proxies for network architecture search},
  author={Lee, Junghyup and Ham, Bumsub},
  booktitle={Proceedings of the IEEE/CVF conference on computer vision and pattern recognition},
  pages={5893--5903},
  year={2024}
}

@inproceedings{tan2019mnasnet,
  title={Mnasnet: Platform-aware neural architecture search for mobile},
  author={Tan, Mingxing and Chen, Bo and Pang, Ruoming and Vasudevan, Vijay and Sandler, Mark and Howard, Andrew and Le, Quoc V},
  booktitle={Proceedings of the IEEE/CVF conference on computer vision and pattern recognition},
  pages={2820--2828},
  year={2019}
}

@inproceedings{cai2019once,
  title={Once-for-all: Train one network and specialize it for efficient deployment},
  author={Cai, Han and Gan, Chuang and Wang, Tianzhe and Zhang, Zhekai and Han, Song},
  booktitle={International Conference on Learning Representations},
  year={2020}
}

@inproceedings{chen2021bn,
  title={Bn-nas: Neural architecture search with batch normalization},
  author={Chen, Boyu and Li, Peixia and Li, Baopu and Lin, Chen and Li, Chuming and Sun, Ming and Yan, Junjie and Ouyang, Wanli},
  booktitle={Proceedings of the IEEE/CVF international conference on computer vision},
  pages={307--316},
  year={2021}
}

@inproceedings{zoph2018learning,
  title={Learning transferable architectures for scalable image recognition},
  author={Zoph, Barret and Vasudevan, Vijay and Shlens, Jonathon and Le, Quoc V},
  booktitle={Proceedings of the IEEE conference on computer vision and pattern recognition},
  pages={8697--8710},
  year={2018}
}

@inproceedings{yang2020cars,
  title={Cars: Continuous evolution for efficient neural architecture search},
  author={Yang, Zhaohui and Wang, Yunhe and Chen, Xinghao and Shi, Boxin and Xu, Chao and Xu, Chunjing and Tian, Qi and Xu, Chang},
  booktitle={Proceedings of the IEEE/CVF conference on computer vision and pattern recognition},
  pages={1829--1838},
  year={2020}
}

@inproceedings{moons2021distilling,
  title={Distilling optimal neural networks: Rapid search in diverse spaces},
  author={Moons, Bert and Noorzad, Parham and Skliar, Andrii and Mariani, Giovanni and Mehta, Dushyant and Lott, Chris and Blankevoort, Tijmen},
  booktitle={Proceedings of the IEEE/CVF International Conference on Computer Vision},
  pages={12229--12238},
  year={2021}
}

@inproceedings{xu2019pc,
  title={Pc-darts: Partial channel connections for memory-efficient architecture search},
  author={Xu, Yuhui and Xie, Lingxi and Zhang, Xiaopeng and Chen, Xin and Qi, Guo-Jun and Tian, Qi and Xiong, Hongkai},
  booktitle={International Conference on Learning Representations},
  year={2020}
}

@inproceedings{yu2020bignas,
  title={Bignas: Scaling up neural architecture search with big single-stage models},
  author={Yu, Jiahui and Jin, Pengchong and Liu, Hanxiao and Bender, Gabriel and Kindermans, Pieter-Jan and Tan, Mingxing and Huang, Thomas and Song, Xiaodan and Pang, Ruoming and Le, Quoc},
  booktitle={European Conference on Computer Vision},
  pages={702--717},
  year={2020},
  organization={Springer}
}

@inproceedings{yang2021towards,
  title={Towards improving the consistency, efficiency, and flexibility of differentiable neural architecture search},
  author={Yang, Yibo and You, Shan and Li, Hongyang and Wang, Fei and Qian, Chen and Lin, Zhouchen},
  booktitle={Proceedings of the IEEE/CVF conference on computer vision and pattern recognition},
  pages={6667--6676},
  year={2021}
}

@inproceedings{xu2022analyzing,
  title={Analyzing and mitigating interference in neural architecture search},
  author={Xu, Jin and Tan, Xu and Song, Kaitao and Luo, Renqian and Leng, Yichong and Qin, Tao and Liu, Tie-Yan and Li, Jian},
  booktitle={International Conference on Machine Learning},
  pages={24646--24662},
  year={2022},
  organization={PMLR}
}

@article{luo2020semi,
  title={Semi-supervised neural architecture search},
  author={Luo, Renqian and Tan, Xu and Wang, Rui and Qin, Tao and Chen, Enhong and Liu, Tie-Yan},
  journal={Advances in Neural Information Processing Systems},
  volume={33},
  pages={10547--10557},
  year={2020}
}

@article{hundt2019sharpdarts,
  title={sharpdarts: Faster and more accurate differentiable architecture search},
  author={Hundt, Andrew and Jain, Varun and Hager, Gregory D},
  journal={arXiv preprint arXiv:1903.09900},
  year={2019}
}

@inproceedings{deng2009imagenet,
  title={Imagenet: A large-scale hierarchical image database},
  author={Deng, Jia and Dong, Wei and Socher, Richard and Li, Li-Jia and Li, Kai and Fei-Fei, Li},
  booktitle={2009 IEEE conference on computer vision and pattern recognition},
  pages={248--255},
  year={2009},
  organization={Ieee}
}

@inproceedings{sandler2018mobilenetv2,
  title={Mobilenetv2: Inverted residuals and linear bottlenecks},
  author={Sandler, Mark and Howard, Andrew and Zhu, Menglong and Zhmoginov, Andrey and Chen, Liang-Chieh},
  booktitle={Proceedings of the IEEE conference on computer vision and pattern recognition},
  pages={4510--4520},
  year={2018}
}

@article{liu2021survey,
  title={A survey on evolutionary neural architecture search},
  author={Liu, Yuqiao and Sun, Yanan and Xue, Bing and Zhang, Mengjie and Yen, Gary G and Tan, Kay Chen},
  journal={IEEE transactions on neural networks and learning systems},
  volume={34},
  number={2},
  pages={550--570},
  year={2021},
  publisher={IEEE}
}

@inproceedings{shu2021nasi,
  title={NASI: Label-and data-agnostic neural architecture search at initialization},
  author={Shu, Yao and Cai, Shaofeng and Dai, Zhongxiang and Ooi, Beng Chin and Low, Bryan Kian Hsiang},
  booktitle={International Conference on Learning Representations},
  year={2022}
}

@article{ning2021evaluating,
  title={Evaluating efficient performance estimators of neural architectures},
  author={Ning, Xuefei and Tang, Changcheng and Li, Wenshuo and Zhou, Zixuan and Liang, Shuang and Yang, Huazhong and Wang, Yu},
  journal={Advances in Neural Information Processing Systems},
  volume={34},
  pages={12265--12277},
  year={2021}
}

@inproceedings{cai2018proxylessnas,
  title={Proxylessnas: Direct neural architecture search on target task and hardware},
  author={Cai, Han and Zhu, Ligeng and Han, Song},
  booktitle={International Conference on Learning Representations},
  year={2019}
}

@inproceedings{hu2022generalizing,
  title={Generalizing few-shot NAS with gradient matching},
  author={Hu, Shoukang and Wang, Ruochen and Hong, Lanqing and Li, Zhenguo and Hsieh, Cho-Jui and Feng, Jiashi},
  booktitle={International Conference on Learning Representations},
  year={2022}
}

@article{zheng2025information,
  title={An information theory-inspired strategy for automated network pruning},
  author={Zheng, Xiawu and Ma, Yuexiao and Xi, Teng and Zhang, Gang and Ding, Errui and Li, Yuchao and Chen, Jie and Tian, Yonghong and Ji, Rongrong},
  journal={International Journal of Computer Vision},
  volume={133},
  number={8},
  pages={5455--5482},
  year={2025},
  publisher={Springer}
}

@inproceedings{ma2024outlier,
  title={Outlier-aware slicing for post-training quantization in vision transformer},
  author={Ma, Yuexiao and Li, Huixia and Zheng, Xiawu and Ling, Feng and Xiao, Xuefeng and Wang, Rui and Wen, Shilei and Chao, Fei and Ji, Rongrong},
  booktitle={Forty-first International Conference on Machine Learning},
  year={2024}
}

@inproceedings{ma2023solving,
  title={Solving oscillation problem in post-training quantization through a theoretical perspective},
  author={Ma, Yuexiao and Li, Huixia and Zheng, Xiawu and Xiao, Xuefeng and Wang, Rui and Wen, Shilei and Pan, Xin and Chao, Fei and Ji, Rongrong},
  booktitle={Proceedings of the IEEE/CVF Conference on Computer Vision and Pattern Recognition},
  pages={7950--7959},
  year={2023}
}
